\pdfoutput=1

\documentclass[11pt]{article}

\usepackage[final]{acl}

\usepackage{times}
\usepackage{latexsym}
\usepackage{color}
\usepackage{graphicx}
\usepackage{multirow}
\usepackage{makecell}
\usepackage{subfigure}
\usepackage{algorithm}
\usepackage{algpseudocode}
\usepackage{booktabs} 
\usepackage{amssymb}
\usepackage{amsmath}
\usepackage{bbm}
\usepackage{xspace}
\usepackage{pifont}
\usepackage[table]{xcolor}

\usepackage[T1]{fontenc}

\usepackage[utf8]{inputenc}

\usepackage{microtype}

\usepackage{inconsolata}

\usepackage{graphicx}

\title{Mitigating Catastrophic Forgetting in Large Language Models with Forgetting-aware Pruning}

\author{
 \textbf{Wei Huang\thanks{These authors contributed equally to this work}},
 \textbf{Anda Cheng$^*$},
 \textbf{Yinggui Wang}\thanks{Corresponding author (wyinggui@gmail.com).},
\\
 Ant Group, China
\\
   hw378176@antgroup.com, andacheng.cad@gmail.com,wyinggui@gmail.com
}

\begin{document}
\maketitle
\begin{abstract}
Recent advancements in large language models (LLMs) have shown impressive capabilities in various downstream tasks but typically face Catastrophic Forgetting (CF) during fine-tuning.
In this paper, we propose the Forgetting-Aware Pruning Metric (FAPM), a novel pruning-based approach to balance CF and downstream task performance. Our investigation reveals that the degree to which task vectors (i.e., the subtraction of pre-trained weights from the weights fine-tuned on downstream tasks)
overlap with pre-trained model parameters is a critical factor for CF. Based on this finding, FAPM employs the ratio of the task vector to pre-trained model parameters as a metric to quantify CF, integrating this measure into the pruning criteria. Importantly, FAPM does not necessitate modifications to the training process or model architecture, nor does it require any auxiliary data.
We conducted extensive experiments across eight datasets, covering natural language inference, General Q\&A, Medical Q\&A, Math Q\&A, reading comprehension, and cloze tests. 
The results demonstrate that FAPM limits CF to just 0.25\% while maintaining 99.67\% accuracy on downstream tasks.
We provide the code to reproduce our results.\footnote{https://github.com/secretflow/ACoLab/tree/main/\\PaperCode/FAPM}.

\end{abstract}

\section{Introduction}

LLMs have demonstrated impressive general capabilities in handling various tasks ~\citep{bubeck2023sparks, rafailov2024direct}. 
Nevertheless, practical deployment frequently uncovers the necessity for augmenting domain-specific competencies~\citep{scialom2022fine}. 
To this end, task-oriented datasets are harnessed to fine-tune these models, enhancing their efficacy in targeted downstream tasks ~\citep{yang2024zhongjing}.
Previous studies have found that while LLMs acquire specialized knowledge from fine-tuning, they tend to forget their general capabilities, especially in full parameter fine-tuning, which is also known as Catastrophic Forgetting (CF)~\citep{luo2023empirical,kong2023overcoming,wumitigating}. 
Consequently, devising methods to alleviate CF during the fine-tuning phase has become a critical research direction for LLMs.

\begin{figure}[t]
\centering
  \includegraphics[width=1.0\linewidth]{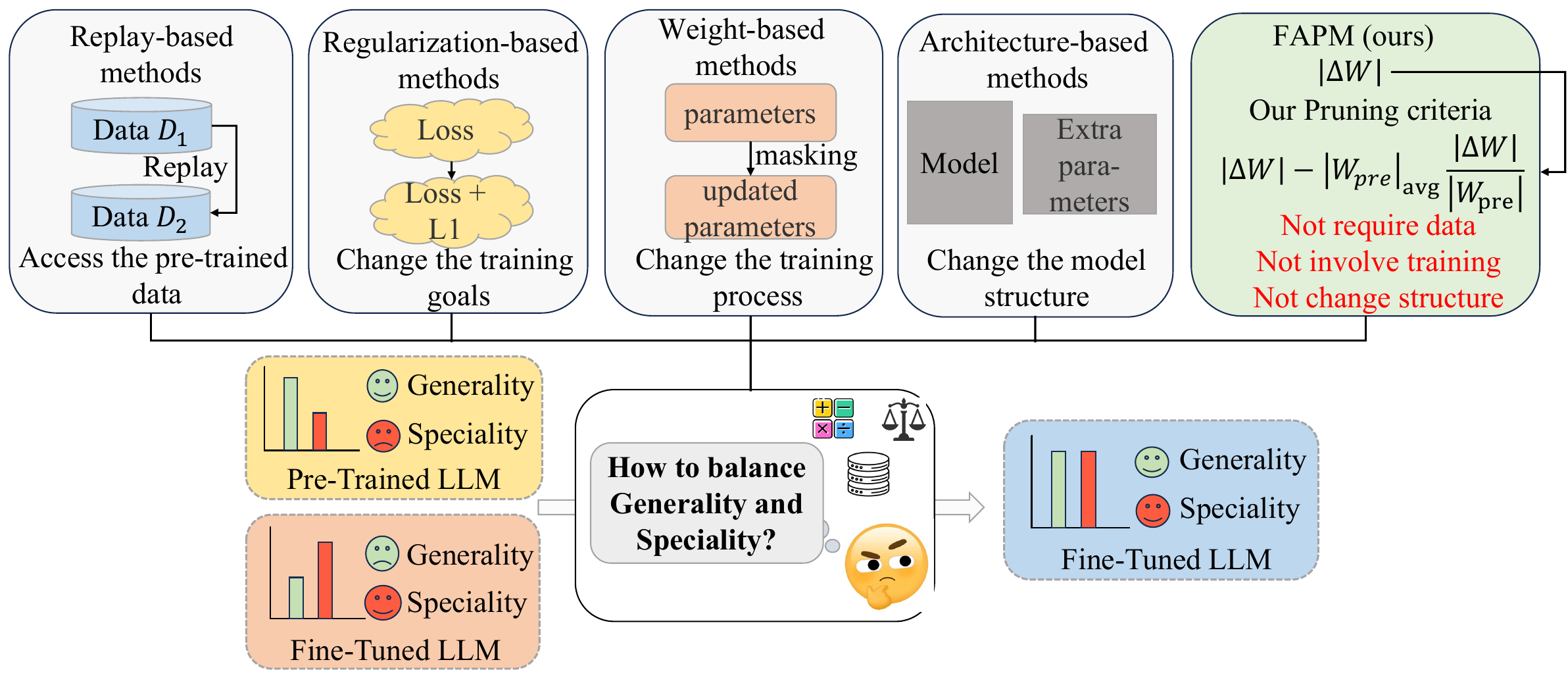}
  \caption {Illustration of the issues in CF, the desired objectives, and the methods to mitigate CF.}
  \label{fig1}
  \vspace{-0.5cm}
\end{figure}

Existing methods to mitigate CF can be divided into four categories: 
1) Replay-based methods incorporate a portion of the pre-training data into the fine-tuning data for training~\citep{scialom2022fine,huang2024mitigating}. 
2) Regularization-based methods introduce additional penalty terms in the loss function, encouraging the fine-tuned model to remain close to the pre-trained model~\citep{lin2023speciality,panigrahi2023task}. 
3) Weight-based methods introduce parameter weight coefficients to regulate their updates~\citep{kecontinual,zhang2024balancing}. 
4) Architecture-based methods design additional modules outside of the original model~\citep{wang2023rehearsal,hu2021lora}. 
Although these methods can alleviate the forgetting problem to a certain extent, they still have the following limitations:
1) The methods that assume a certain amount of pre-training data can be obtained are unrealistic in practical applications because many open-source LLMs have not released their pre-training data.
2) Even if pre-training data could be obtained, incorporating it into the fine-tuning process would significantly increase training costs.
\citep{kecontinual,zhang2024balancing}. 
3) Methods that alter the training process or model architecture make the training process more difficult to control. Moreover, in practice, adapting to downstream tasks often requires adjusting different hyperparameters of the method. Since this involves fine-tuning, it can also introduce significant time and computational costs.

These limitations raise the following question:

\emph{Can we solve the problem of catastrophic forgetting \textbf{without changing training process}, \textbf{without any additional data}, and \textbf{without altering model structure}? 
} 

Recent research has highlighted two key findings: 
1) There are a significant number of redundant parameters in LLMs~\citep{yadav2024ties}. 
2) The task vector specifies a direction in the weight space of the pre-trained model and moving towards its direction can improve task performance~\citep{ilharcoediting}. 
These findings suggest that we can prune portions of the task vector's parameters and set them to zero. 
By doing so, the corresponding positions of the pre-trained model's parameters are exposed, potentially preserving the accuracy of downstream tasks while mitigating CF to some extent.
To this end, we first try to apply existing pruning methods to prune the task vector to alleviate CF. 
Unfortunately, we find it challenging to strike an optimal balance between maintaining downstream task accuracy and mitigating CF using existing pruning techniques alone~\citep{han2015learning,wanda}. 
Specifically, pruning the task vector with a low sparsity ratio fails to effectively mitigate CF, whereas pruning with a high sparsity ratio results in poor downstream task accuracy. 
We find that there are two main reasons for this problem:
1) The existing pruning criteria only ensure the balance between downstream task accuracy and sparsity, while not considering CF. 
2) The extent to which the values of task vectors overlap with pre-trained model parameters is a critical factor contributing to CF. 

In this paper, we propose a new pruning method called Forgetting-Aware Pruning Metric (FAPM). FAPM not only applies magnitude as the pruning criterion for task vectors but also uses the ratio of task vectors to pre-trained model parameters as the criterion for mitigating CF. 
By adopting FAPM, we aim to identify, during the pruning process, those parameters in the task vector whose values are large (crucial for maintaining the accuracy of downstream tasks) and we concurrently intend to penalize those parameters where the ratio of their magnitude in the task vector to that of the corresponding parameter in the pre-trained model is notably high (more likely to induce CF). 
This balanced approach aspires to surgically retain the most valuable parameters for task performance while excising those that pose the greatest risk to the model's generality. Overall, \textbf{FAPM does not require any modifications to the training process or model architecture, nor does it need additional data.} Furthermore, the implementation of FAPM is straightforward and efficient, allowing it to be completed in a very short time.

Extensive experiments on different LLMs and various datasets show that \textbf{FAPM can maintain a downstream task accuracy of up to 99.67\% while the degree of CF is only 0.25\%.} 
Compared to structure-based strategies, such as LoRA, FAPM achieves superiority in precision and forgetting rate. 
\textbf{Our experiments primarily focus on full fine-tuning; however, it is surprising that when LoRA fine-tuning leads to CF, FAPM can also effectively mitigate this issue.} In the paper, we primarily explore mitigating the phenomenon of forgetting in pre-trained models during fine-tuning. Further experiments demonstrate that FAPM continues to be effective in alleviating forgetting during sequential fine-tuning. 

\section{Analysis of Catastrophic Forgetting}

\begin{figure*}[!t]
    \centering
    \subfigure[The original accuracy on RTE is 0.890 and the original average accuracy on  general tasks is 0.6204.]{
        \includegraphics[width=.41\textwidth]{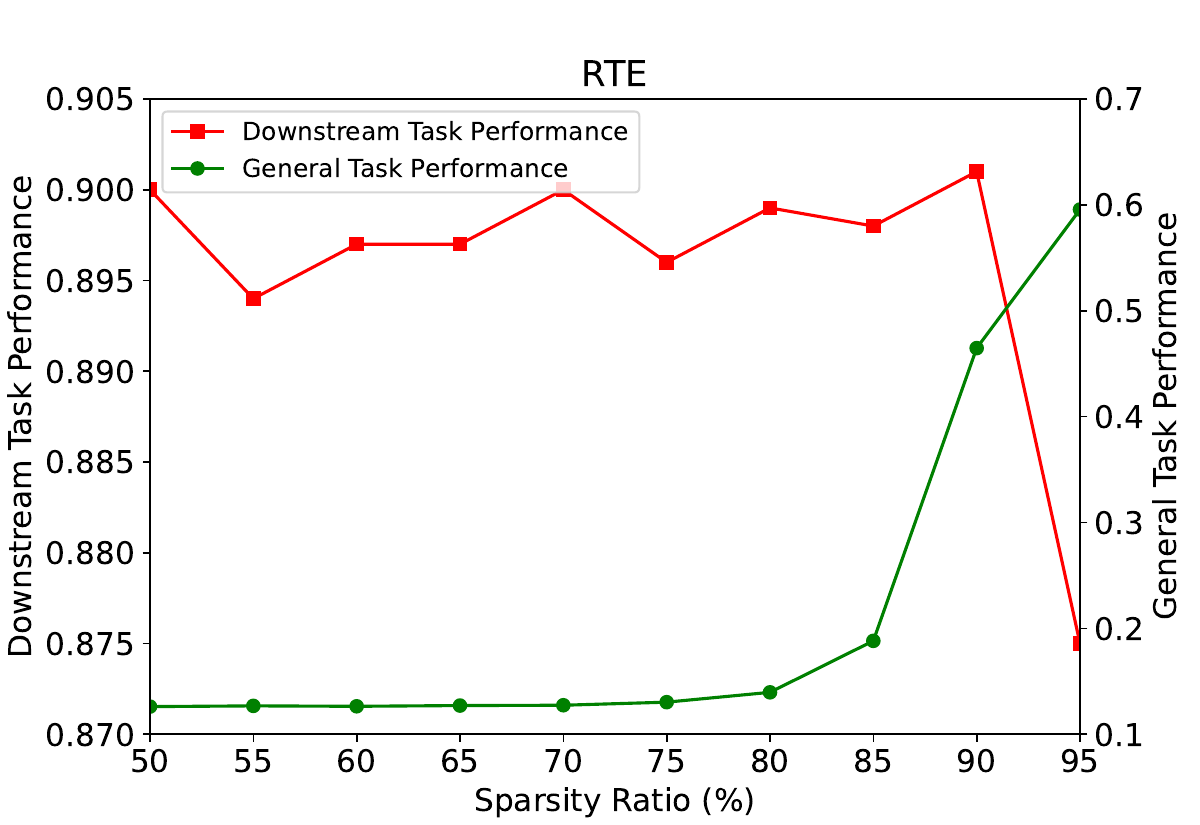}
    }
    \hspace{1cm}
    \subfigure[The original accuracy on MRPC is 0.887 and the original average accuracy on  general tasks is 0.6204.]{
        \includegraphics[width=.41\textwidth]{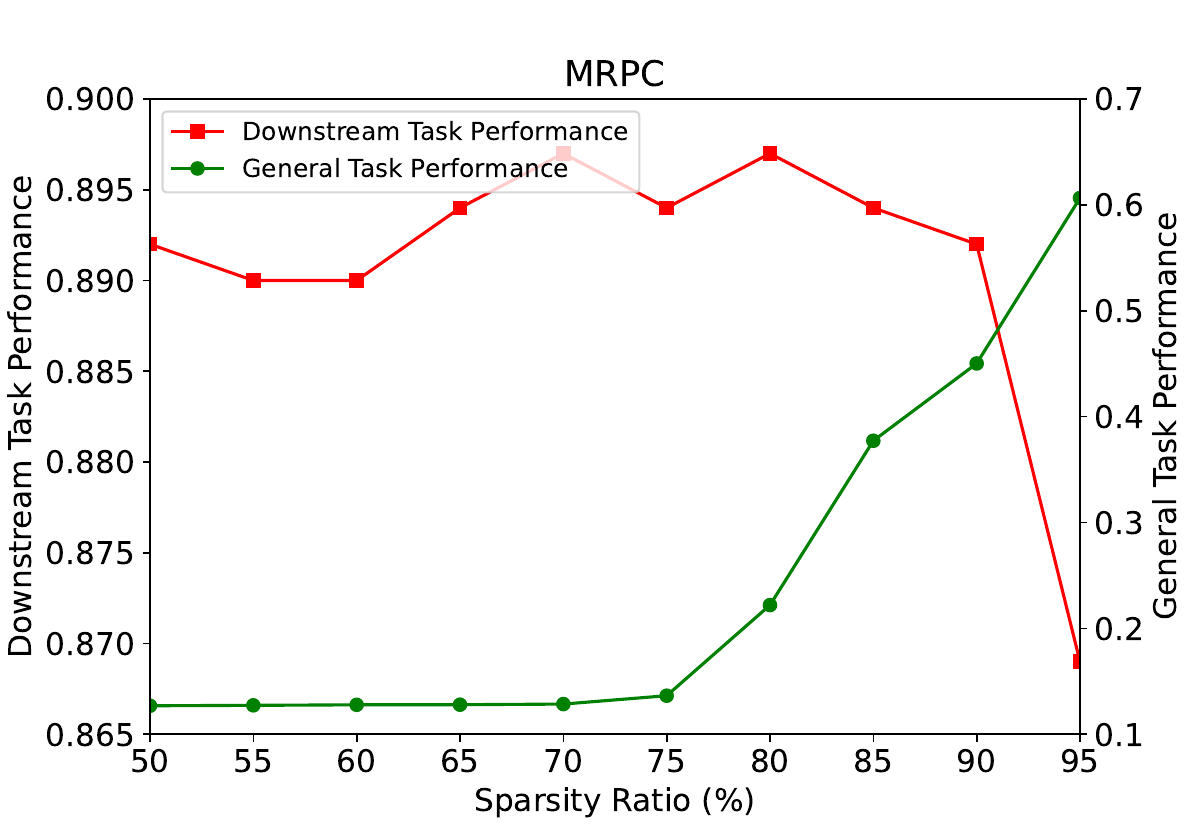}
    }
    \caption {
    The relationship between the magnitude pruning sparsity ratio, general capability, and downstream task performance of Llama3-8B on (a) RTE and (b) MRPC, respectively. 
    When sparsity is below 90\%, downstream task performance remains relatively stable, but CF is notably serious (general task performance is poor). When sparsity exceeds 90\%, increasing sparsity alleviates CF effectively, but significantly reduces downstream task performance. 
    We also observed the same phenomenon on Qwen2-7B, see Appendix ~\ref{appendix3}.
}
\vspace{-0.3cm}
    \label{fig2}
\end{figure*}

\subsection{Background}

\textbf{Problem Setting.} Given downstream data $D$ and a pre-trained model, we fine-tune the model using 
$D$. Let the pre-trained model parameters be $W_{pre}$ and the fine-tuned model parameters be $W_{ft}$. In this paper, we perform a series of operations on the task vector. 
Following previous work~\citep{ilharcoediting}, the task vector $\Delta W \in \mathbb{R}_{d}$ can be defined as $W_{ft} - W_{pre}$. This operation allows us to focus on the changes that occur during the fine-tuning stage.

\textbf{Pruning on the task vector to mitigate CF.}
We begin by mitigating CF using existing pruning methods to prune the task vector according to their magnitude~\citep{han2015learning}. 
The red lines in Figure \ref{fig2} show how performance changes with different sparsity ratios on the RTE and MRPC datasets. At each sparsity level, only the top-k\% highest-magnitude values are retained, and others are set to zero. Green lines illustrate the impact of sparsity on general task performance, particularly CF, where higher accuracy means less forgetting. 
Results indicate many task vector values are redundant, with task accuracy unaffected even at 90\% sparsity. This implies that when the pruned $\Delta W$ is combined with $W_{pre}$, 90\% of $W_{pre}$ parameters are utilized. However, downstream performance deteriorates beyond 90\% sparsity while CF is further reduced. This highlights the challenge in balancing task performance and CF reduction with magnitude pruning of $\Delta W$ and raises a pertinent question:

\emph{What additional factors, beyond $\Delta W$ itself, could also affect the balance between maintaining downstream task accuracy and mitigating CF? }





\subsection{Exploration and Analysis}

\paragraph{Theoretical analysis.} To answer the above question, we begin by analyzing the factors that cause CF. 
Considering that the finetuned model can be expressed as $W_{ft} = W_{pre} + \Delta W$, previous works~\cite{wise-ft, kirkpatrick2017overcoming, panda2024lottery} mainly focus on $\Delta W$ when analyzing CF, and typically mitigate CF by constraining or modifying $\Delta W$.
However, these works often overlook the role of $W_{{pre}}$ and the interplay between $W_{pre}$ and $\Delta W$. Given that the final model is a combination of both components, we argue that their relationship plays a critical role in determining the trade-off between task performance and knowledge retention.
Consider a simple but illustrative case: if $\Delta W = 0$, then $W_{{ft}} = W_{{pre}}$, resulting in no CF, since the model remains unchanged from its pre-trained state. 
Conversely, large deviations from $W_{{pre}}$ are more likely to disrupt previously acquired knowledge. Therefore, the magnitude of $\Delta W$ relative to $W_{{pre}}$ significantly affects the extent of CF. 


We further decompose the fine-tuned model weights as $W_{ft} = W_{pre} + \Delta W = W_{pre} \cdot \left(1 + \frac{\Delta W}{W_{pre}} \right)$, 
the term $1 + \frac{\Delta W}{W_{pre}}$ can be interpreted as a coefficient matrix that reflects the relative change in each parameter of $W_{pre}$ induced during the fine-tuning process.
As $W_{pre}$ itself can not cause CF, this decomposition highlights that CF may be more closely related to the last term $\frac{\Delta W}{W_{{pre}}}$.
Intuitively, for a given parameter, a larger value of $|\frac{\Delta W}{W_{{pre}}}|$ indicates a more significant deviation from the pre-trained state, which is more likely to disrupt previously acquired knowledge. We refer to this quantity as the \emph{relative change magnitude}.
In contrast to the commonly used \emph{absolute change magnitude} criterion (which considers only $|\Delta W|$), the relative change magnitude explicitly accounts for the interplay between $\Delta W$ and $W_{{pre}}$. 
Therefore, modeling this relationship for CF enables a more principled identification of parameters that are most influential in preserving pre-training performance.

\begin{figure*}[t]
\centering
  \includegraphics[width=.88\linewidth]{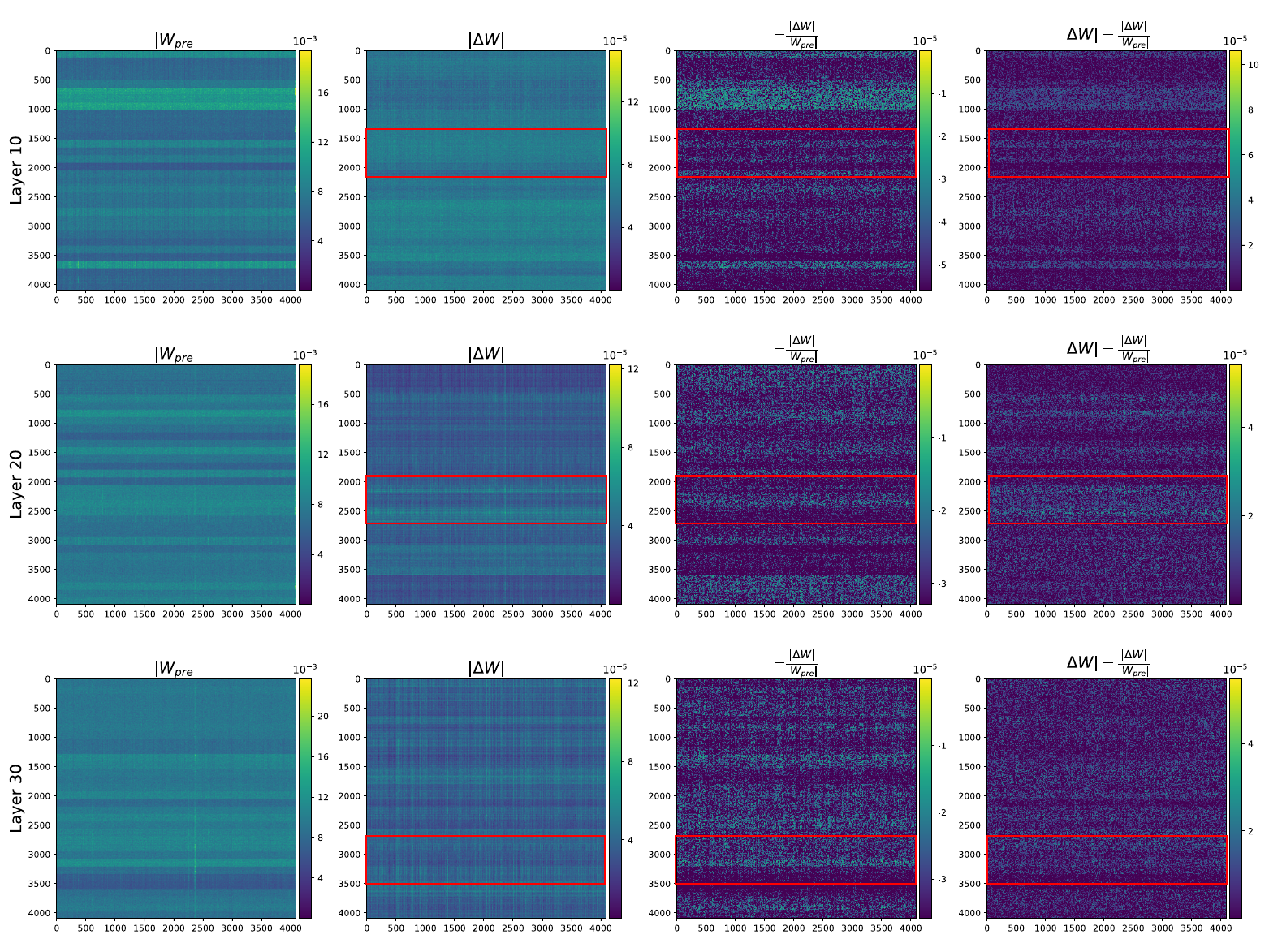}
  \caption {
   Visualization of the weight matrices in different layers of Llama3-8B fine-tuned on RTE dataset. From left to right, they represent the magnitude of the pre-trained model weights, the absolute change magnitude of model weights, the relative change magnitude of model weights, and a combination of the absolute and relative change magnitude. The absolute and relative changes patterns show clear differences, such as the channels marked by the red boxes. 
}
  \label{fig3}
  \vspace{-0.3cm}
\end{figure*}
\paragraph{Empirical analysis.} In Figure \ref{fig3}, we illustrate the differences in attention to various positions within the model weight matrices across different layers, guided by the absolute change magnitude criterion and the relative change magnitude criterion. 
Brighter regions in the figure represent parameters with higher scores under a specific criterion, while darker regions denote parameters with lower scores. 
Under the absolute change magnitude criterion, the highlighted areas indicate parameters crucial for downstream task accuracy. 
In contrast, under the relative change magnitude criterion, the highlighted regions indicate parameters important for mitigating forgetting.

By comparing the images in the middle two columns, we observe a significant divergence in scoring patterns between the absolute change magnitude criterion and the relative change magnitude criterion. 
The highlighted areas under the absolute change magnitude criterion do not entirely correspond to those under the relative change magnitude criterion. 
This discrepancy indicates that parameters retained under the absolute change magnitude criterion may not be effective in mitigating catastrophic forgetting. 
This also explains why it is difficult to balance downstream task accuracy and forgetting when using $|\Delta W|$ as the pruning criterion alone. 
To achieve a more favorable balance between downstream task accuracy and CF, we propose a hybrid pruning criterion that incorporates both the absolute change magnitude and relative change magnitude. 
This combined criterion fuses the strengths of both individual criteria and exhibits a distinct pattern that differs from using either criterion in isolation as shown in the rightmost part of Figure \ref{fig3}.

\section{FAPM: Forgetting-Aware Pruning Metric}
\label{sec3}

\begin{figure*}[!t]
\centering
  \includegraphics[width=.72\linewidth]{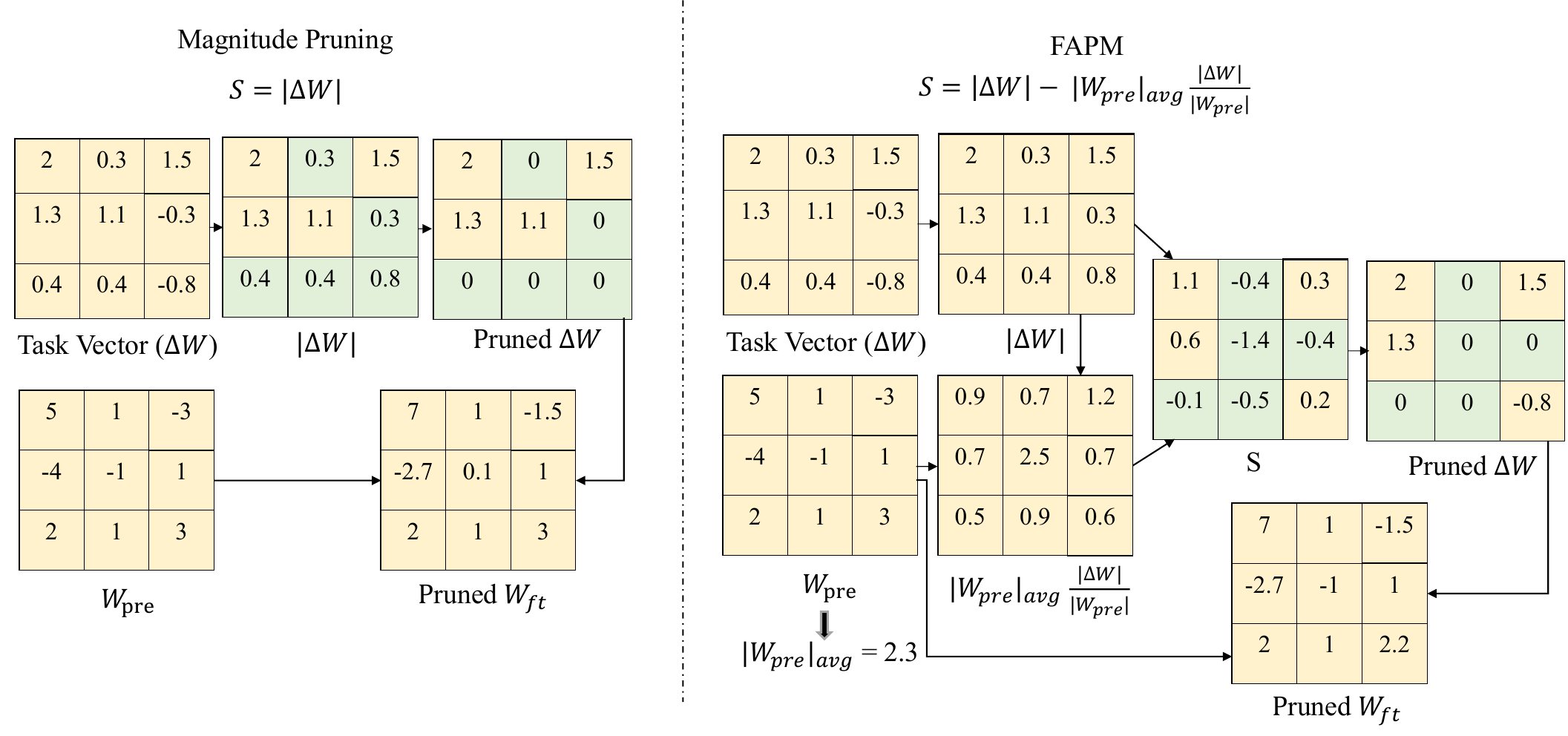}
  \vspace{-0.25cm}
  \caption {Illustration of pruning with FAPM.
  If $|\Delta W^{i}|$ is large (it will be retained by magnitude pruning) and $\frac{|\Delta W^{i}|}{|W_{pre}^{i}|}$ is also large, FAPM will penalize and possibly prune this parameter, e.g., the value $1.1$ in the middle of $\Delta W$. 
  By doing so, most large-magnitude parameters in $\Delta W$ are retained, while only a small subset are replaced by parameters with smaller $\frac{|\Delta W^{i}|}{|W_{pre}^{i}|}$ values.
}
  \vspace{-0.35cm}
  \label{fig4}
\end{figure*}
Consider a linear layer in $\Delta W$ with weights $\Delta W^{i}$ of shape ($C_{in}$, $C_{out}$), corresponding to the linear layer representation $W_{pre}^{i}$ in $W_{pre}$. We propose to evaluate each weight matrix's importance by subtracting the relative change magnitude criterion from the absolute change magnitude criterion, as shown in Figure~\ref{fig4}. 
Our pruning criterion for $\Delta W^{i}$ is formulated as:
\begin{align}\label{eq1}
    S_{i} = |\Delta W^{i}| - |W_{pre}^{i}|_{avg}\frac{|\Delta W^{i}|}{|W_{pre}^{i}|}   
\end{align}
where $\left| \cdot \right|$ denotes the absolute value operation, and $avg$ represents the averaging operation on the parameter matrix. $i$ denotes one of the matrices in the $\Delta W$ parameter matrix. 
We included $|W_{pre}^{i}|_{avg}$ in the formula due to our observations during practical operations. We found that the numerical values of $\frac{|\Delta W^{i}|}{|W_{pre}^{i}|}$ and $|\Delta W^{i}|$ do not fall within the same range. For instance, the order of magnitude of $\frac{|\Delta W^{i}|}{|W_{pre}^{i}|}$ is approximately $1 \times 10^{-2}$, whereas that of $|\Delta W^{i}|$ is around $1 \times 10^{-4}$.
This will lead one criterion to predominate over the other, weakening the impact of the other. Therefore, to balance the numerical values of $\frac{|\Delta W^{i}|}{|W_{pre}^{i}|}$ and $|\Delta W^{i}|$, we have introduced $|W_{pre}^{i}|_{avg}$. 

\begin{table*}[!th]
\centering
\scalebox{0.63}
{\begin{tabular}{cccccccc|ccccccccc}
\toprule
\multicolumn{1}{c}{
\multirow{2}{*}{Tasks}}
& 
\multicolumn{1}{c}{
\multirow{2}{*}{Methods}}
&
\multicolumn{5}{c}{Llama3-8B}
& &
\multicolumn{5}{c}{Qwen2-7B}
\\
\cmidrule{3-14} &
\multicolumn{1}{l}{} 
& C-Eval & GSM8K & MMLU & HumanEval &  Avg. & Results & C-Eval & GSM8K & MMLU & HumanEval &  Avg. & Results 
\\
\hline
\multirow{7}{*}{RTE} & Pre-trained  &0.4386	&0.7922	&0.6594	&0.5914	&0.6204	&0.819 &0.7478	&0.8180	&0.6884	&0.7682	&0.7556	&0.574\\
 & Full SFT &0.2311	&0.075	&0.2554	&0.0	&0.1403	&0.890 &0.2602	&0.075	&0.2423	&0.0	&0.1443	&0.890 \\
 \cmidrule{2-14}
 \cmidrule{2-14} & L1-reg
 &0.3735	&0.7353	&0.6012	&0.5367	&0.5616	&0.843 &0.7108	&0.7463	&0.6143	&0.7118	&0.6958	&0.847  \\
 & WiSE-FT & 
 0.3046	 &0.4420	 &0.4255	 &0.5609	 &0.4332	 & \underline{0.889} &0.7488&	0.7664&	0.4638	&0.7682	&0.6868	&0.888
 \\
 & V-SoftMask
 &0.4144	&0.7811	&0.5702	&0.4919	&0.5644	& {0.886}  &0.7317	&0.7371	&0.6448	&0.7111	&0.7062	& \underline{0.896}\\
 & CoFiTune
 &0.4542	&0.7869	&0.6492	&0.5815	&0.6180	&0.882 &0.7591	&0.8125	&0.6808	&0.7560	& \bf{0.7521}	&0.886 \\
 & LoRA
 &0.4435	&0.7892	&0.6574	& 0.5915	& \underline{0.6204}	&0.866  &0.7456	&0.8133	&0.6897	&0.7500	&0.7496	&0.877 \\
 \rowcolor{gray!10} & FAPM (Ours) &0.4623	&0.7915	&0.6454	&0.5975	&\bf{0.6242}	& \bf{0.897} &0.7568	&0.8104	&0.6857	&0.7500	& \underline{0.7507}	& \bf{0.903}  \\
\hline
\hline
\multirow{7}{*}{WikiQA} & Pre-trained  &0.4386	&0.7922	&0.6594	&0.5914	&0.6204	&0.913 &0.7478	&0.8180	&0.6884	&0.7682	&0.7556	& 0.896 \\
 & Full SFT &0.2547	&0.0	&0.2422	&0.0	&0.1242	&0.966 &0.2510	&0.076	&0.2434	&0.0	&0.1426	&0.965\\
 \cmidrule{2-14}
 \cmidrule{2-14} & L1-reg
 &0.4271	&0.7591	&0.5780	&0.5549	&0.5797	&0.945  &0.6818	&0.7582	&0.6186	&0.7091	&0.6919	&0.955 \\
& WiSE-FT &  0.2581	 &0.0	 &0.2458	 &0.0 &	0.1259	 &0.958  &0.2476 &	0.0	 &0.2414 &	0.0	 &0.1222	 &0.961 \\
 & V-SoftMask
 &0.2944	&0.7282	&0.5677	&0.2910	&0.4703	& \underline{0.963}  &0.6862	&0.6585	&0.5331	&0.6759	&0.6384	& 0.962\\
 & CoFiTune
 &0.4164	&0.7702	&0.6309	&0.5666	&0.5960	&0.960  &0.7527	&0.7755	&0.6358	&0.7195	&0.7208	&\underline{0.961} \\
 & LoRA
 &0.4423	&0.8013	&0.6429	&0.5919	& \underline{0.6196}	&0.955   &0.7519	&0.8019	&0.6873	&0.7621	& \underline{0.7507}	&0.960\\
 \rowcolor{gray!10} & FAPM (Ours) &0.4749	&0.7975	&0.6563	&0.5853	& \bf{0.6285}	& \bf{0.964} &  0.7555	&0.8036	&0.6902	&0.7621	& \bf{0.7529}	& \bf{0.962}  \\
\hline
\hline
\multirow{7}{*}{Winogrande} & Pre-trained  &0.4386	&0.7922	&0.6594	&0.5914	&0.6204	&0.519&0.7478	&0.8180	&0.6884	&0.7682	&0.7556	&0.558 \\
 & Full SFT &0.2792	&0.0606	&0.3438	&0.0	&0.1709	&0.820&0.4090	&0.0303	&0.2996	&0.0609	&0.1999	&0.790 \\
 \cmidrule{2-14} \cmidrule{2-14} & L1-reg
 &0.4234	&0.7572	&0.6245	&0.5667	&0.5904	&0.737  &0.7283	&0.7609	&0.6401	&0.7277	&0.7143	&0.703 \\
 &WiSE-FT	&0.4741&	0.4412	&0.5967	&0.5060	&0.5045	& \bf{0.830} & 0.6747&	0.5343	&0.5821	&0.5914	&0.5956	&0.780
 \\
 & V-SoftMask
 &0.4089	&0.7017	&0.5528	&0.5003	&0.5409	& {0.828} &0.7321	&0.7098	&0.6241	&0.6861	&0.6880	& \bf{0.791} \\
 & CoFiTune
 &0.4719	&0.7817	&0.6410	&0.5743	&0.6172	&0.813  &0.7550	&0.7990	&0.6820	&0.7500	&\underline{0.7465}	&0.771\\
 & LoRA
 &0.4522	&0.7822	&0.6429	&0.5775	& \underline{0.6137}	&0.810 &0.7430	&0.8118	&0.6761	&0.7500	& 0.7452	&0.782 \\
 \rowcolor{gray!10} & FAPM (Ours)  &0.4829	&0.7680	&0.6472	&0.5731	& \bf{0.6178}	& \underline{0.824} &0.7618	&0.8068	&0.6845	&0.7395	& \bf{0.7482}	& \underline{0.785}   \\
\hline
\hline
\multirow{7}{*}{SQuAD} & Pre-trained  &0.4386	&0.7922	&0.6594	&0.5914	&0.6204	&0.371&0.7478	&0.8180	&0.6884	&0.7682	&0.7556	&0.451 \\
 & Full SFT &0.2806	&0.0212	&0.3206	&0.0	&0.1556	&0.646  &0.3531	&0.0212	&0.3183	&0.0	&0.1731	&0.624\\
 \cmidrule{2-14} & L1-reg 
 &0.3990	&0.6605	&0.5800	&0.5113	&0.5377	&0.565  &0.6481	&0.6614	&0.5883	&0.6681	&0.6414	&0.561\\
 &WiSE-FT	&0.4309	&0.4009	&0.5484	&0.5102	&0.4725	&0.639 &0.6784	&0.5743	&0.5868	&0.5524	&0.5979	&0.622
 \\
 & V-SoftMask
 &0.3757	&0.0786	&0.4755	&0.5013	&0.3578	&0.635  &0.6369	&0.5881	&0.5933	&0.6451	&0.6159	& \bf{0.624}\\
 & CoFiTune
 &0.4319	&0.7596	&0.6356	&0.5766	& \underline{0.6009} &0.633   &0.7451	&0.7626	&0.6584	&0.7621	& \underline{0.7321}	&0.619 \\
 & LoRA
 &0.4795	&0.7255	&0.5914	&0.5853	&0.5954	& \bf{0.648} &0.7253	&0.7665	&0.6537	&0.7482	&0.7234	& \underline{0.620} \\
 \rowcolor{gray!10} & FAPM (Ours) &0.4738	&0.7310	&0.6455	&0.5748	& \bf{0.6063}	& \underline{0.637}  &0.7410	&0.8006	&0.6752	&0.7500	& \bf{0.7417}	&0.615 \\
 \hline
\hline
\hline
\multirow{7}{*}{MedQA} & Pre-trained  & 0.4386 & 0.7922 & 0.6594 & 0.5914 & 0.6204 & 0.552  & 0.7478 & 0.818 & 0.6884 & 0.7682 & 0.7556 & 0.531 \\
 & Full SFT  & 0.3315 & 0.4897 & 0.5365 & 0.4434 & 0.4502 & 0.636  & 0.6920 & 0.7293 & 0.6343 & 0.6463 & 0.6754 & 0.613 \\
 \cmidrule{2-14} & L1-reg & 0.4040 & 0.7511 & 0.6464 & 0.5291 &0.5826 & 0.606  & 0.6938 & 0.7802 & 0.6837 & 0.6955 &0.7133  & 0.567\\
 &WiSE-FT & 0.3976 & 0.6847 & 0.6049 & 0.4628 &0.5375 & 0.647  & 0.6248 & 0.7083 & 0.6493 & 0.6111 &0.6483  & \underline{0.607}\\
 & V-SoftMask & 0.3940 & 0.7287 & 0.6048 & 0.5279 &0.5638 & 0.641  & 0.6738 & 0.7802 & 0.6474 & 0.6805 &0.6954  & 0.606
\\
 & CoFiTune &0.4338 &0.7560 &0.6575& 0.5639&\underline{0.6028} &0.640 &0.7488	&0.8191	&0.6971	&0.7656	&\underline{0.7576}	&0.604
 \\
 & LoRA
 &0.4216 &0.7089 &0.5889& 0.5409& 0.5650 &\underline{0.640} &0.7072	&0.7527	&0.6404	&0.7256	&0.7064	&0.589
 \\
 \rowcolor{gray!10} & FAPM (Ours)  & 0.4586 & 0.7733 & 0.6638 & 0.5731 & \bf{0.6172} & \bf{0.643}  & 0.7346 & 0.8241 & 0.6935 & 0.7412 & \underline{0.7484} & \bf{0.613} \\
\bottomrule
\end{tabular}}
\caption{Results on various datasets on Llama3-8B and Qwen2-7B models with full parameter Fine-tuning. "Avg." means the average results across the C-Eval, GSM8K, MMLU, HumanEval datasets. Results indicate the accuracy on the different downstream task datasets. 
 The bold and underlined data represent the best and second-best results, respectively. The experimental results for the other three datasets can be found in Appendix ~\ref{sero}.}
\label{table1}
\vspace{-.5cm}
\end{table*}

Our FAPM has several intriguing properties. Firstly, when the value of a parameter in 
$|\Delta W^{i}|$ is large (indicating that the parameter would typically be retained according to traditional magnitude pruning criteria) and simultaneously, $\frac{|\Delta W^{i}|}{|W_{pre}^{i}|}$ is also large (suggesting that this parameter may contribute to catastrophic forgetting), the FAPM pruning strategy will penalize and potentially prune this parameter. 
Under the FAPM criteria, to ensure downstream task accuracy, most parameters with large magnitudes will still be retained, while only a small subset will be replaced by parameters with smaller $\frac{|\Delta W^{i}|}{|W_{pre}^{i}|}$ values.
Secondly, the computation of FAPM is both simple and efficient. We only need to obtain the fine-tuned and pre-trained model parameters, eliminating the need for additional data. The computational overhead associated with this method is minimal, enhancing the generalizability of FAPM. 
We provide the pseudocode implementation of FAPM in Appendix \ref{appendix1}.

\section{Experiment Setup}
\textbf{Datasets and Setting:} We evaluate FAPM on Llama3-8B~\citep{dubey2024llama} and Qwen2-7B~\citep{yang2024qwen2}. 
Following prior studies~\citep{yadav2024ties,wu2024reft,han2024parameter}, we evaluate models' specialized performance across four tasks: natural language inference, question answering, cloze tests, and reading comprehension. 
We utilize the MRPC
~\citep{wang2019glue} and RTE
~\citep{wang2019glue} datasets for natural language inference, with accuracy as the evaluation metric. 
For question answering, we employ the WikiQA
~\citep{yang-etal-2015-wikiqa}, QASC
~\citep{allenai:qasc}, MedQA
~\citep{jin2021disease}, and MetaMathQA
~\citep{yu2023metamath}. For the first three datasets, we used ROUGE-L as the evaluation metric, and for MetaMathQA, we used GSM8K~\citep{cobbe2021gsm8k} as the evaluation dataset. 
We use the Winogrande dataset
~\citep{sakaguchi2021winogrande} for cloze tests, measuring performance with accuracy. 
Lastly, we utilize the SQuAD dataset
~\citep{rajpurkar-etal-2016-squad} for reading comprehension, with the F1-score as the evaluation metric.
To evaluate the generality of LLMs, we integrate insights from previous studies~\citep{dubey2024llama,yang2024qwen2} and focus on four key aspects.
We use MMLU~\citep{hendryckstest2021} to assess the inherent world knowledge stored in the LLM, C-Eval~\citep{huang2023ceval} to evaluate the model's understanding of general knowledge in Chinese, GSM8K~\citep{cobbe2021gsm8k} to evaluate mathematical reasoning, and HumanEval~\citep{chen2021evaluating} to assess the code generation capabilities. The setup of our experimental process can be found in Appendix ~\ref{es}.

\textbf{Compared Methods} We compared FAPM with the full-parameter SFT (Full SFT) and five CF baselines, which are described in detail in Appendix ~\ref{appendix2}. These baselines are carefully categorized into three groups: 1) Regularization-based: 
The chosen baseline for comparison is the L1 regularization~\citep{kirkpatrick2017overcoming}. 2) Weight-based: 
The selected baselines include V-SoftMask~\citep{kecontinual},WiSE-FT~\citep{wise-ft}, and CoFiTune~\citep{zhang2024balancing}. 3) Architecture-based: 
The baseline under this category is LoRA~\citep{hu2021lora}.


\begin{table*}[!t]
\centering
\scalebox{0.63}
{\begin{tabular}{cccccccc|ccccccccc}
\toprule
\multicolumn{1}{c}{
\multirow{2}{*}{Tasks}}
& 
\multicolumn{1}{c}{
\multirow{2}{*}{Pruning Rate}}
&
\multicolumn{5}{c}{Llama3-8B}
& &
\multicolumn{5}{c}{Qwen2-7B}
\\
\cmidrule{3-14} & \multicolumn{1}{l}{}
& C-Eval & GSM8K & MMLU & HumanEval &  Avg. & Results & C-Eval & GSM8K & MMLU & HumanEval &  Avg. & Results 
\\
\hline
\multirow{4}{*}{SQuAD} & $0\%$  & 0.4795 & 0.7255 & 0.5914 & 0.5853 & 0.5954 & 0.648 & 0.7253  & 0.7665  & 0.6537  & 0.7482  & 0.7234 & 0.620 \\
& $3\%$   & 0.4591 & 0.7801 & 0.6511 & 0.5914 & 0.6204 & 0.644 & 0.7359 & 0.8225 & 0.6852 & 0.7621 & 0.7514 & 0.621 \\
& $10\%$   & 0.4502 & 0.7862 & 0.6552 & 0.5914 & 0.6207 & 0.616 & 0.7466 & 0.8195 & 0.6869 & 0.7560 & 0.7522 & 0.618 \\
& $90\%$   & 0.4435 & 0.8013 & 0.6594 & 0.5853 & 0.6223 & 0.456 & 0.7447 & 0.8180 & 0.6884 & 0.7517 & 0.7507 & 0.579 \\
\hline
\multirow{4}{*}{MedQA} & $0\%$ &0.4216 &0.7089 &0.5889& 0.5409& 0.5650 &0.640 &0.7072	&0.7527	&0.6404	&0.7256	&0.7064	&0.589  \\
& $3\%$ &0.4592 &0.7807	&0.6484	&0.5731	&0.6154	&0.632 &0.7478	&0.8111	&0.6776	&0.7651	&0.7504	&0.589 \\
& $10\%$&0.4540	&0.7807	&0.6439	&0.5875	&0.6165	&0.623 &0.7588	&0.8101	&0.6824	&0.7661	&0.7544	&0.571 \\
& $90\%$&0.4544	&0.7899	&0.6590	&0.5914	&0.6236	&0.588 &0.7472	&0.8137	&0.6804	&0.7756	&0.7542	&0.557 \\
\bottomrule
\end{tabular}}
\caption{Results of FAPM on various datasets and different models using LoRA Fine-tuning.}
\label{table2}
\end{table*}

\begin{table*}[!t]
\centering
\scalebox{0.73}{%
\begin{tabular}{ccccc}
\toprule
\multirow{2}{*}{Sparsity Ratios} & \multicolumn{4}{c}{Dataset} \\ 
\cmidrule{2-5} 
& Winogrande (Magnitude) & Winogrande ($|\Delta W|/|W_{pre}|$) & WikiQA (Magnitude) & WikiQA ($|\Delta W|/|W_{pre}|$) \\ 
\hline
$90\%$  & 0.6178	&0.5936	&0.6206	&0.6014 \\ 
$80\%$  & 0.6119	&0.5542	&0.6135	&0.5569 \\ 
$70\%$ & 0.5953	&0.5001	&0.6007	&0.5387 \\ 
$60\%$ & 0.5882	&0.4267	&0.5735	&0.4477 \\ 
$50\%$ & 0.5768	&0.3548	&0.5345	&0.4089 \\
\bottomrule
\end{tabular}}
\caption{Results of the relationship between relative change
magnitude and CF. We pruned $|\Delta W|$ by removing portions with small values of $|\Delta W|/|W_{pre}|$ and compared the results with magnitude-based pruning. The above results are the average accuracy across four general datasets.}
\label{table_rcm}
\vspace{-.4cm}
\end{table*}

\section{Results}

In this section, we aim to investigate the effectiveness of the FAPM. 
The evaluation focuses on downstream tasks' performance and generalization ability metrics, using the performance of Full SFT and the pre-trained models as reference points. 

\emph{Due to space limitations in the main paper, we present two additional experiments in the appendix:}

\emph{1) We explore the performance of FAPM under different sparsity ratios. The experimental results are shown in Appendix ~\ref{app-abl}.}

\emph{2) We investigate the relationship between various unstructured pruning criteria and CF to explain why FAPM is improved based on Magnitude Pruning rather than the SOTA LLM pruning methods. The results can be found in Appendix ~\ref{pc}.}

\subsection{Comparison with Various Baselines}

In Tables \ref{table1}, we present the comparative results of FAPM and various baselines on different datasets. 
The results indicate that Full SFT exhibits significant forgetting on Llama3-8B, with average accuracy on general datasets maintaining only around 0.15.
This indicates that the fine-tuned model loses almost all generalization capability, demonstrating that full fine-tuning severely exacerbates CF. 
On the Llama3-8B model, FAPM achieves an average performance of 0.814 on eight downstream datasets, compared to Full SFT's 0.816, a decrease of only 0.25\%, indicating that FAPM has minimal impact on downstream task performance. 
Furthermore, FAPM's average performance on the general tasks is 0.6112 (0.6132 for the Pre-trained model), a decrease of only 0.33\% compared to the Pre-trained model, demonstrating that FAPM significantly alleviates CF.
Similar trends are observed with the Qwen2-7B model. 

Compared to L1-regularization, FAPM demonstrates a stronger ability to preserve downstream task accuracy and better addresses CF. 
Specifically, for the Llama3-8B model, L1-regularization results in an average performance drop of 5.99\% across eight downstream datasets. 
While both LoRA and FAPM similarly mitigate CF, LoRA slightly compromises downstream task accuracy, particularly on the MRPC and RTE datasets. 
V-SoftMask excels in preserving downstream task accuracy but performs poorly in addressing CF, with an average performance drop of 10.92\% on  general tasks. 
Compared to the CoFiTune method, FAPM also demonstrates comparable performance. 
Overall, FAPM shows strong performance when compared to existing regularization-based, weight-based, and architecture-based methods.

\subsection{Effectiveness for LoRA Fine-tuning}
According to Table \ref{table1}, we observed a trend of CF when applying LoRA fine-tuning on the MedQA and SQuAD datasets. We conducted further experiments to validate the effectiveness of FAPM in LoRA fine-tuning.
The original FAPM defines the task vector as $W_{ft} - W_{pre}$. 
In LoRA fine-tuning, the initialization of LoRA weights is not based on the pre-trained model weights. 
Therefore, the original definition of task vectors cannot be directly applied to LoRA fine-tuning.
Here we edit the original FAPM to fit LoRA fine-tuning.
In LoRA fine-tuning, we can merge LoRA parameters with the pre-trained model parameters as $W_{new} = W_{pre} + W_{loraB}W_{loraA}$, where $W_{loraB}$ and $W_{loraA}$ are LoRA matrices. 
Merging the LoRA parameters with the pre-trained model parameters involves the dot product of the $W_{loraB}$ and $W_{loraA}$ matrices and then addition to the pre-trained model weights. Therefore, we can treat $W_{loraB}W_{loraA}$ as the task vectors in our algorithm for pruning. That is, $\Delta W$ in Eq.\ref{eq1} is $W_{loraB}W_{loraA}$. From Table \ref{table2}, we can see that 
FAPM effectively alleviates CF to a mere 0.6\% while maintaining an impressive 99.5\% accuracy on downstream tasks in both datasets. 
Table \ref{table2} also reveals that the pruning ratio in LoRA fine-tuning needs to be relatively small to avoid significant performance degradation in downstream tasks. We speculate that this is mainly because the LoRA matrices $W_{loraB}W_{loraA}$ are low-rank, and thus contain relatively few redundant parameters.

\subsection{Exploration relationship between relative change magnitude and CF} 

In Table ~\ref{table_rcm}, we conducted experiments to analyze the relationship between $|\Delta W|/|W_{pre}|$ and CF. We pruned $|\Delta W|$ by removing portions with small values of $|\Delta W|/|W_{pre}|$ and compared the results with magnitude-based pruning. Experiments were performed on Llama3-8B using the Winogrande and WikiQA datasets. 
When retaining the same proportion of parameters, the pruning strategy based on $|\Delta W|/|W_{pre}|$ is more likely to cause CF, indicating that using $|\Delta W|/|W_{pre}|$ to evaluate different parameters is more effective in identifying those that are subject to forgetting. This reveals a strong relationship between $|\Delta W|/|W_{pre}|$ and CF. When the sparsity rate is set at 50\%, the magnitude-based pruning performance remains around 0.55, while $|\Delta W|/|W_{pre}|$-based pruning approximately 0.4.

\subsection{Effectiveness for Sequential Fine-tuning}
In the paper, we primarily explore the ability of pre-trained models to forget after fine-tuning. However, readers may also be concerned about whether FAPM is effective in multi-task sequential training. The phenomenon of forgetting in multi-task sequential training refers to the loss of learned capabilities from prior tasks while learning the current one during sequential task training. For example, if a model is fine-tuned using dataset A and then further fine-tuned using dataset B, it may forget the capabilities learned from dataset A. To demonstrate the effectiveness of FAPM in multi-task sequential training, we conducted experiments using three datasets: RTE, WikiQA, and Winogrande. The results are presented in Table ~\ref{table:st} and ~\ref{table:st1}. The "Original" column shows the model accuracy for each sequential training session, while "CF" refers to the accuracy of each task after multi-task sequential training, which can lead to catastrophic forgetting (CF). The findings indicate that our method remains effective in multi-task sequential training.

\begin{table}[t]
    \centering
\scalebox{0.75}{%
\begin{tabular}{lccc}
\toprule
Datasets & Original & CF & FAPM (Ours) \\ 
\midrule
RTE     & 0.890    & 0.652 & 0.878 \\ 
WikiQA  & 0.961    & 0.961 & 0.958 \\ 
\bottomrule
\end{tabular}}
\caption{The effectiveness of FAPM in sequential training. The model used is Llama3-8B, with the order of tasks being RTE, WikiQA.}
\vspace{-0.3cm}
\label{table:st}
\end{table}

\begin{table}[t]
    \centering
\scalebox{0.75}{%
\begin{tabular}{lccc}
\toprule
Datasets & Original & CF & FAPM (Ours) \\ 
\midrule
RTE     & 0.890    & 0.716 & 0.882 \\ 
Winogrande      & 0.812    & 0.812 & 0.805 \\ 
\bottomrule
\end{tabular}}
\caption{The effectiveness of FAPM in sequential training with the order of tasks being RTE, Winogrande.}
\label{table:st1}
\vspace{-0.3cm}
\end{table}

\section{Related Work}
\paragraph{Catastrophic Forgetting in LLMs.}
Fine-tuning LLMs, a common practice to enhance model specialization, often leads to CF. ~\citep{luo2023empirical}. 
Existing approaches to mitigate CF can be broadly categorized into four main categories:
1) Replay-based methods ~\citep{huang2024mitigating} typically integrate some pre-training data into the fine-tuning dataset for training.
2) Regularization-based methods ~\citep{lin2023speciality} introduce additional penalty terms in the loss function, encouraging the fine-tuned model to maintain proximity to the pre-trained model.
3) Weight-based methods ~\citep{zhang2024balancing} introduce parameter weight coefficients to modulate their updates, thereby ensuring controlled adjustments during the optimization process. 
4) Architecture-based methods ~\citep{wang2023rehearsal,hu2021lora} involve the design of additional modules external to the original model.

\paragraph{LLM Pruning.}
Network pruning \citep{NIPS1989_6c9882bb} is considered a popular approach for compressing LLMs, which shrinks model sizes by removing specific weights.
Magnitude Pruning \citep{han2015learning} 
removes individual weights based on their magnitudes, where weights with magnitudes
below a certain threshold are removed. 
Recent LLM pruning methods typically involve calculating pruning metrics according to model weights and activations by using some additional data. 
Wanda \citep{wanda} prunes weights with the
smallest magnitudes multiplied by the norm of
the corresponding input activations, without the
need for retraining or weight updates.
All these methods aim to reduce the model parameters while maintaining model performance.
Different from this, in this paper, we intend to achieve a better balance mitigating CF and improving downstream accuracy by pruning task vectors in LLM fine-tuning.

\section{Conclusion}

In this study, we present a straightforward and efficient method to tackle the issue of CF that emerges during the continuous fine-tuning of LLMs. 
We find that the extent to which task vectors overlap with the pre-trained model parameters is a key factor influencing CF. 
Based on this observation, we propose FAPM to effectively address CF while preserving the performance of the fine-tuning tasks. 
FAPM integrates the ratio of the task vector to the pre-trained model parameters as a criterion, combining it with the magnitude-based pruning metric.  
It does not require any additional auxiliary data, nor does it necessitate alterations to the training process or model structure. 

\section*{Limitations}

While FAPM demonstrates promising performance, we did not investigate its integration with existing CF techniques to address the problem of forgetting. Specifically, our FAPM is a post-processing method, whereas most existing methods for mitigating CF are training-based. In theory, these two approaches can be combined; however, determining how to maximize the synergistic effects of such a combination remains an important direction for future research.

\bibliography{custom}

\appendix

\begin{table*}[!th]
\centering
\scalebox{0.63}
{\begin{tabular}{cccccccc|ccccccccc}
\toprule
\multicolumn{1}{c}{
\multirow{2}{*}{Tasks}}
& 
\multicolumn{1}{c}{
\multirow{2}{*}{Methods}}
&
\multicolumn{5}{c}{Llama3-8B}
& &
\multicolumn{5}{c}{Qwen2-7B}
\\
\cmidrule{3-14} &
\multicolumn{1}{l}{} 
& C-Eval & GSM8K & MMLU & HumanEval &  Avg. & Results & C-Eval & GSM8K & MMLU & HumanEval &  Avg. & Results 
\\
\hline
\multirow{7}{*}{MathQA} & Pre-trained   & 0.4386 & - & 0.6594 & 0.5914 & 0.5631 & 0.792  & 0.7478 & - & 0.6884 & 0.7682 & 0.7347 & 0.818\\
 & Full SFT  & 0.3101 & - & 0.4411 & 0.1890 & 0.3133 & 0.818 & 0.5528 & - & 0.6056 & 0.1890 & 0.4491 & 0.828 \\
 \cmidrule{2-14} & L1-reg & 0.3974 & - & 0.5955 & 0.5360 &0.5096  & 0.788 & 0.7043 & - & 0.6217 & 0.7102 & 0.6787 & 0.806\\
  &WiSE-FT & 0.3648 & - & 0.5476 & 0.4609 &0.4577  & 0.808 & 0.6865 & - & 0.5887 & 0.7030 &0.6594  & 0.842\\
 & V-SoftMask & 0.3963 & - & 0.5715 & 0.5483 &0.5053  & \underline{0.810} & 0.6835 & - & 0.6171 & 0.6766 &0.6590  & \underline{0.851}
\\
 & CoFiTune & 0.4449 & - & 0.6409 & 0.5711 &0.5523  & 0.805 & 0.7549 & - & 0.6763 & 0.7492 & \underline{0.7268} & 0.848
 \\
 & LoRA  & 0.4651 & - & 0.6575 & 0.5609 & \underline{0.5611} & 0.809 & 0.7499 & - & 0.6822 & 0.7478 & 0.7266 & 0.826 \\
 \rowcolor{gray!10} & FAPM (Ours)  & 0.4697 & - & 0.6447 & 0.5729 & \bf{0.5624} & \bf{0.812} & 0.7374 & - & 0.6899 & 0.7560 & \bf{0.7278} & \bf{0.851} \\
 \hline
\hline
\multirow{7}{*}{MRPC} & Pre-trained & 0.4386 &0.7922& 0.6594 &0.5914 &0.6204 &0.686 &0.7478 &0.8180& 0.6884& 0.7682 &0.7556 &0.765 \\
 & Full SFT & 0.2603 &0.0& 0.2483& 0.0 &0.1271 &0.887 & 0.2598& 0.0& 0.2481& 0.0 &0.1269 &0.914\\
 \cmidrule{2-14} 
 & L1-reg&0.4062	&0.7470	&0.6200	&0.5434	&0.5766	&0.821 &0.7136	&0.7779	&0.6261	&0.7171	&0.7086	&0.823 \\
 & WiSE-FT &0.3382	&0.7012	&0.5528	&0.4660	&0.5145	&0.878 &0.6564	&0.6119	&0.5551	&0.6027	&0.6065	&0.886\\
 & V-SoftMask &0.4200	&0.7474	&0.5229	&0.5122	&0.5506	&\underline{0.878}  &0.7418	&0.6933	&0.6095	&0.6901	&0.6836	&0.889\\
 & CoFiTune &0.4513	&0.7863	&0.6382	&0.5821	&\underline{0.6145}	&0.874  &0.7612	&0.8036	&0.6795	&0.7317	&0.7440	&\underline{0.889} \\
 & LoRA &0.4546	&0.7890	&0.6506	&0.5936	&\textbf{0.6210}	&0.846  &0.7468	&0.8125	&0.6873	&0.7439	&0.7476	&0.873 \\
 \rowcolor{gray!10} & FAPM (Ours) &0.4662	&0.7711	&0.6410	&0.5791	&0.6144	&\bf{0.882}  &0.7564	&0.7938	&0.6837	&0.7682	&\textbf{0.7505}	&\textbf{0.892} \\
 \hline
\hline
\multirow{7}{*}{QASC} & Pre-trained model &0.4386	&0.7922	&0.6594	&0.5914	&0.6204	&0.630 &0.7478	&0.8180	&0.6884	&0.7682	&0.7556	&0.701\\
 & Full SFT &0.4284	&0.0379	&0.5115	&0.0121	&0.2474	&0.864 &0.5876	&0.0470	&0.5445	&0.2621	&0.3603	&0.866\\
 \cmidrule{2-14} & L1-reg&0.4133	&0.7744	&0.6119	&0.5507	&0.5875	&0.802 &0.7300	&0.7813	&0.6453	&0.7091	&0.7164	&0.781  \\
 & WiSE-FT &0.4332&	0.4766	&0.5367	&0.5009	&0.4868	& \underline{0.858} & 0.6641&	0.5510	&0.5776	&0.5549	&0.5869	&0.851\\
 & V-SoftMask &0.4372	&0.7245	&0.5922	&0.5781	&0.5830	&0.853 &0.7452	&0.7636	&0.6388	&0.7195	&0.7167	&\textbf{0.857} \\
 & CoFiTune &0.4836	&0.7919	&0.6457	&0.5992	&\textbf{0.6301}	&0.835 &0.7744	&0.8006	&0.6778	&0.7500	&\underline{0.7507}	&0.848 \\
 & LoRA &0.4833	&0.7930	&0.6471	&0.5731	&0.6241	&0.856 &0.7677	&0.8218	&0.6872	&0.7134	&0.7475	&\underline{0.855}  \\
 \rowcolor{gray!10} & FAPM (Ours) &0.4836	&0.7983	&0.6326	&0.5914	&\underline{0.6265}	&\textbf{0.863}&0.7679	&0.8157	&0.6815	&0.7500	&\textbf{0.7538}	&0.851    \\
\bottomrule
\end{tabular}}
\caption{Supplementary results on various datasets on Llama3-8B and Qwen2-7B models with full parameter Fine-tuning. "MathQA" refers to "MetaMathQA". Since both MathQA and GSM8K are mathematical datasets, we do not report the GSM8K  performance of the model trained on MathQA.}
\label{table1_app}
\end{table*}

\section{Pseudocode for FAPM}
\label{appendix1}
In this section, we describe the pseudocode for FAPM. A detailed introduction to FAPM can be found in Section ~\ref{sec3} of the main paper.
\begin{algorithm}
\caption{FAPM Procedure}
\label{algo:pseudo-code}
\begin{algorithmic}
\Ensure pre-trained model $W_{pre}$, fine-tuned model $W_{ft}$, layer number $L$, desired sparsity $s$.
\Require pruned $W_{ft}^{i}$.
\For{$i \in [0,L]$}
    \State $\Delta W^{i} = W_{ft}^{i} - W_{pre}^{i}$. 
    \State Calculate score vector $S^{i} \leftarrow |\Delta W^{i}| - \operatorname{Avg}( |W_{pre}^{i}|) * \frac{|\Delta W^{i}|}{|W_{pre}^{i}|}$.
    \State Obtain pruning threshold $t^{i}$ according to $s$ and $S^{i}$.
    \State Obtain pruning mask matrix $ M^{i} = \textbf{1}[\![S^{i}>t^{i}]\!]$.
    \State  $\Delta W^{i} \leftarrow \Delta W^{i} \odot M^{i}$.
    \State $W_{ft}^{i}$ = $W_{pre}^{i}$ + $\Delta W^{i}$. 
\EndFor
\end{algorithmic}
\end{algorithm}

\section{Supplementary Experimental Setup}

\subsection{Experimental Setting} 
\label{es}
During training, we set the learning rate to 1e-5 and the batch size to 64. Each dataset was trained for 3 epochs. The AdamW optimizer was used for fine-tuning. We employed LLaMA-Factory~\citep{zheng2024llamafactory} as the training platform and vLLM~\citep{kwon2023efficient} for inference. When implementing the FAPM algorithm, we applied a 90\% sparsity rate across all models and datasets. All experiments are conducted with 4 NVIDIA A100 GPUs with 80G memory.

\subsection{Baseline Descriptions}
\label{appendix2}
In this Section, we describe the baseline method in our setting in detail.

L1 regularization ~\citep{kirkpatrick2017overcoming} adds an L1 penalty term to the original loss function to promote sparsity in the parameter updates. The modified loss function is $L\left( \theta \right) \  +\  \lambda_{1} \| \theta -\theta_{pre} \|_{1}$, with the regularization hyperparameter set to 1e-6.

Ke et al.~\citep{kecontinual} proposed the Vanilla Soft-masking method to address the issue of catastrophic forgetting in language models during continual fine-tuning. Specifically, this method employs a gradient-based detection technique to calculate the importance of units within both the attention and feed-forward network (FFN) modules across all transformer layers. The obtained importance weights are then used to control the backpropagation of gradients.

Wortsman et al.~\citep{wise-ft} proposed Wise-FT, which consists of two stages: 1) fine-tuning a pre-trained model on the downstream task, and 2) merging the pre-trained and fine-tuned models through linear weight interpolation.

Zhang et al.~\citep{zhang2024balancing} proposed the CoFiTune method to tackle the issue of catastrophic forgetting. CoFiTune employs a two-stage approach. At the coarse-grained level, an empirical tree-search algorithm is used to identify and update specific modules that are crucial for the fine-tuning task, while keeping other parameters frozen. At the fine-grained level, a soft-masking mechanism is employed to adjust the updates of the large model, thereby alleviating catastrophic forgetting.

Inspired by the perspective that ``pre-trained models have a lower intrinsic dimension when fine-tuned on specific tasks," Hu et al.~\citep{hu2021lora} proposed a fine-tuning method called LoRA. During the training process of LoRA, the pre-trained parameters are kept frozen to preserve their general capabilities, while all the decomposition matrices within the low-rank matrix are trainable.

\section{Supplementary experimental results for the other datasets}
\label{sero}

In Tables \ref{table1_app}, we present the comparative results of FAPM and various baselines on the other three datasets. These results indicate that our proposed FAPM method effectively maintains downstream task performance while alleviating CF.

\begin{figure}[!th]
    \centering
    \begin{minipage}[b]{0.49\textwidth}
        \centering
        \includegraphics[width=.80\textwidth]{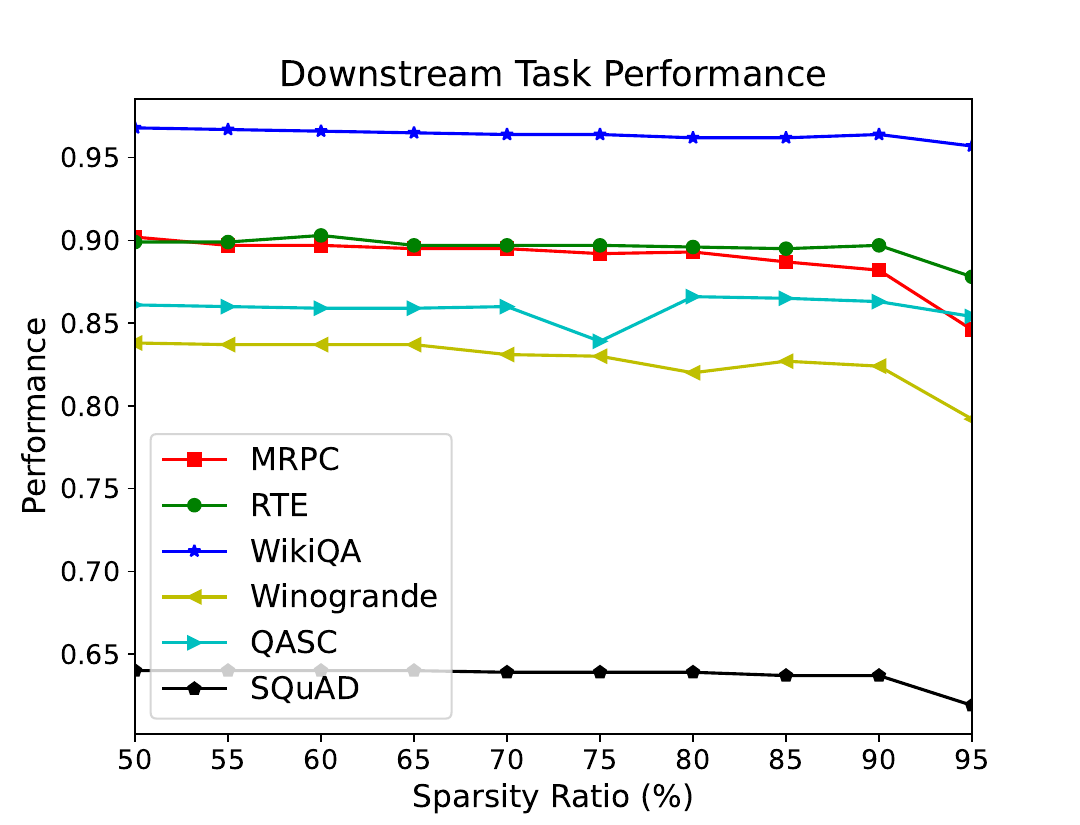}
    \end{minipage}
    \hfill
    \begin{minipage}[b]{0.49\textwidth}
        \centering
        \includegraphics[width=.80\textwidth]{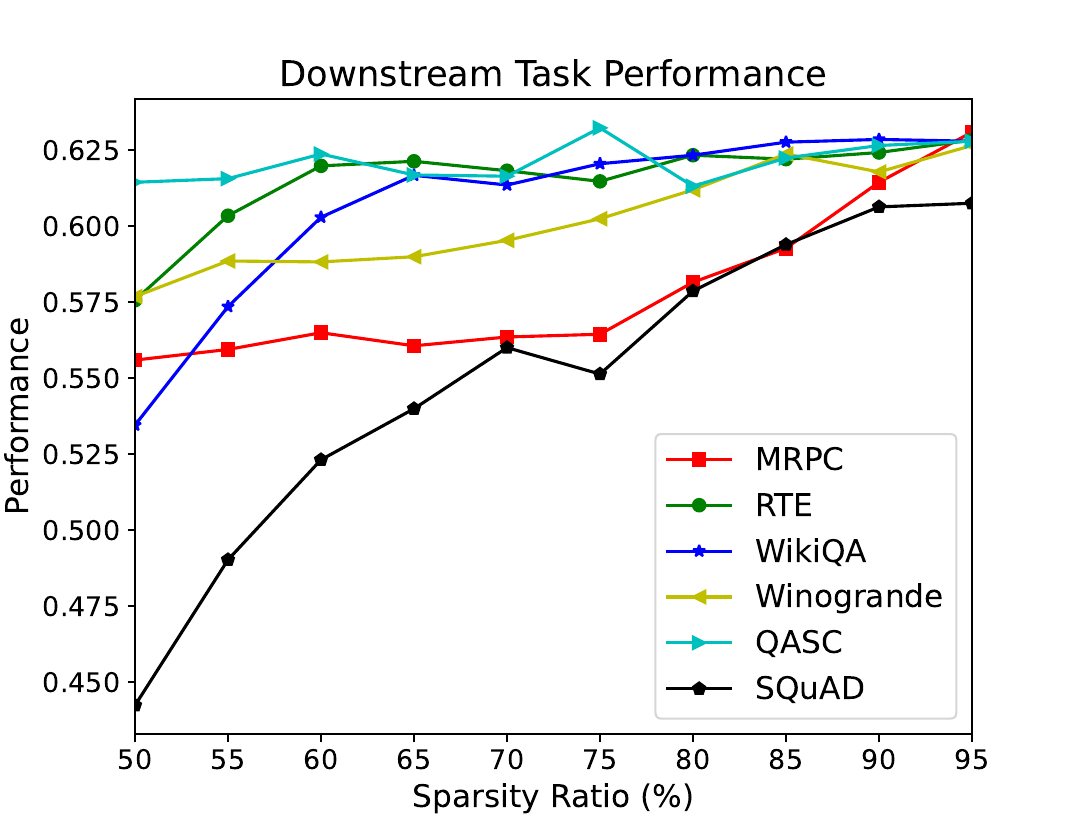}
    \end{minipage}
    \caption {Performance of FAPM on downstream task accuracy and mitigation of catastrophic forgetting with different sparsity ratios on Llama3-8B. }
    \label{fig5}
\end{figure}
\begin{figure}[!th]
    \centering
    \begin{minipage}[b]{0.49\textwidth}
        \centering
        \includegraphics[width=.80\textwidth]{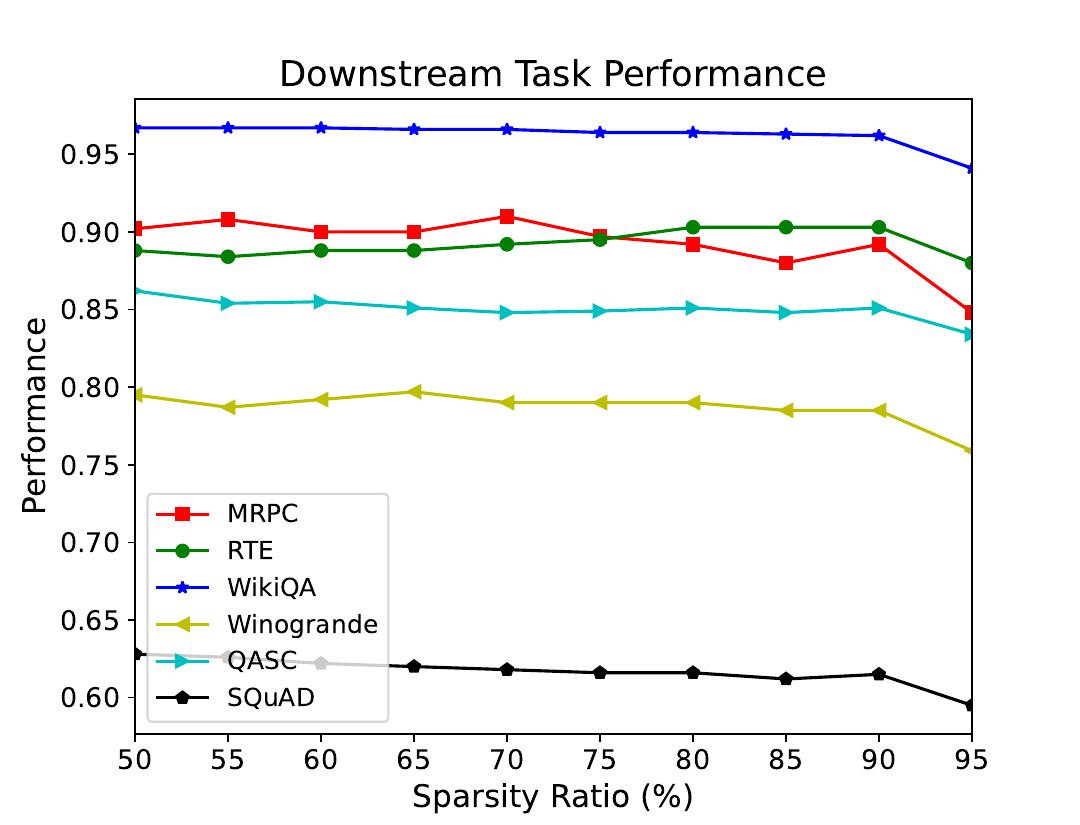}
    \end{minipage}
    \hfill
    \begin{minipage}[b]{0.49\textwidth}
        \centering
        \includegraphics[width=.80\textwidth]{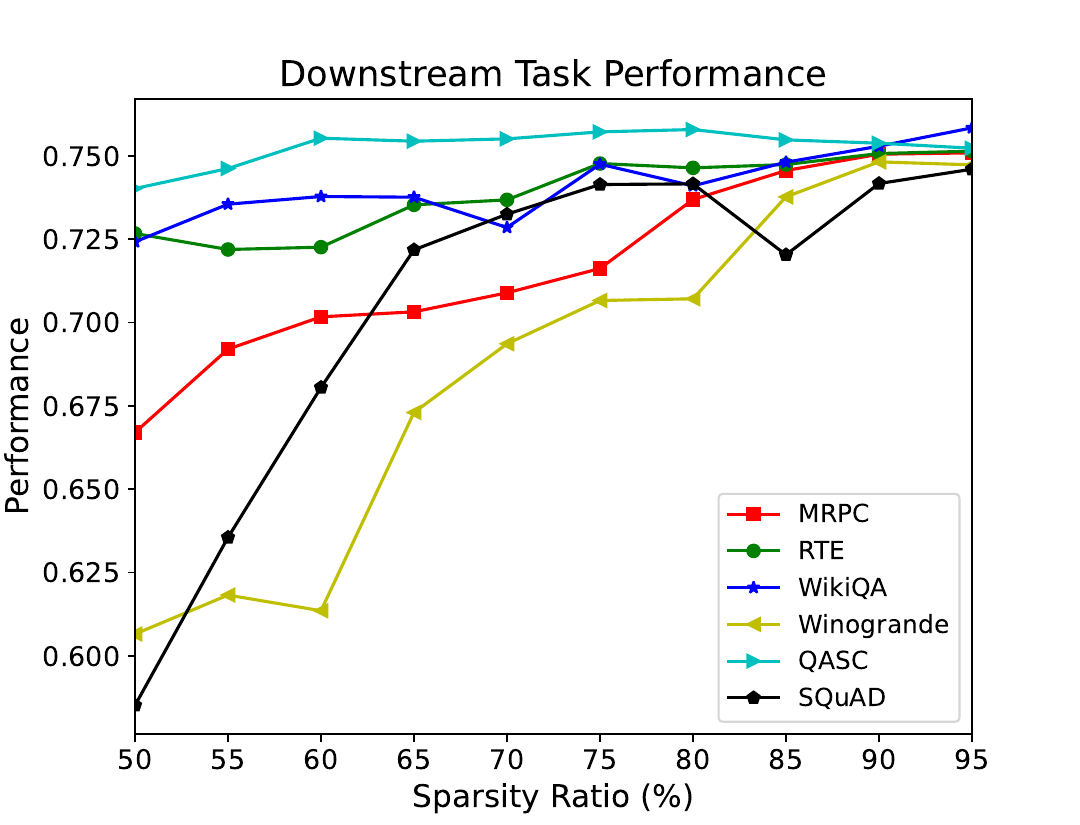}
    \end{minipage}
    \caption {Performance of FAPM on downstream task accuracy and mitigation of catastrophic forgetting with different sparsity ratios on Qwen2-7B. }
    \label{fig6}
\end{figure}

\section{Ablation Studies of Sparsity}\label{app-abl}

\begin{table*}[!t]
\centering
\scalebox{0.63}{\begin{tabular}{cccccccc|ccccccccc}
\toprule
 \multicolumn{1}{c}{
\multirow{2}{*}{Tasks}}
& 
\multicolumn{1}{c}{
\multirow{2}{*}{Methods}}
&
\multicolumn{5}{c}{Llama3-8B}
& &
\multicolumn{5}{c}{Qwen2-7B}
\\
\cmidrule{3-14} &
\multicolumn{1}{l}{} 
& C-Eval & GSM8K & MMLU & HumanEval &  Avg. & Results & C-Eval & GSM8K & MMLU & HumanEval &  Avg. & Results \\
\hline
\multirow{3}{*}{RTE} & Magnitude
&0.3063	&0.6631	&0.4052	&0.4843	&0.4647	&\bf{0.901} &0.7144	&0.7346	&0.5190	&0.6943	&0.6655	& \underline{0.895}  \\
 & Wanda
 &0.4675	&0.7827	&0.6465	&0.5732	& \underline{0.6174}	&0.878 &0.7442	&0.8025	&0.6774	&0.7542	& \underline{0.7445}	&0.877 \\
 \rowcolor{gray!10} & FAPM (Ours) &0.4623	&0.7915	&0.6454	&0.5975	&\bf{0.6242}	& \underline{0.897} &0.7568	&0.8104	&0.6857	&0.7500	& \bf{0.7507}	& \bf{0.903}  \\
\hline
\multirow{3}{*}{WikiQA} & Magnitude
&0.2606	&0.0	&0.2553	&0.0	&0.1289	& \underline{0.964} &0.7162	&0.0416	&0.2560	&0.0	&0.2535	& \bf{0.965} \\
 & Wanda
 &0.4760	&0.7804	&0.6432	&0.5834	& \underline{0.6207}	&0.961  &0.7520	&0.7793	&0.6784	&0.7134	& \underline{0.7307}	&0.958\\
 \rowcolor{gray!10} & FAPM (Ours) &0.4749	&0.7975	&0.6563	&0.5853	& \bf{0.6285}	& \bf{0.964}  &0.7555	&0.8036	&0.6902	&0.7621	& \bf{0.7529}	& \underline{0.962}  \\
\hline
\multirow{3}{*}{Winogrande} & Magnitude
&0.4957	&0.6148	&0.6236	&0.5731	&0.5768	& \bf{0.828}&0.6849	&0.5056	&0.6133	&0.5975	&0.6003	& \underline{0.742}  \\
 & Wanda
 &0.4748	&0.7762	&0.6508	&0.5919	& \bf{0.6234}	&0.750  &0.7549	&0.7915	&0.6806	&0.5914	& \underline{0.7046}	&0.731 \\
 \rowcolor{gray!10} & FAPM (Ours) &0.4829	&0.7680	&0.6472	&0.5731	& \underline{0.6178}	& \underline{0.824} &0.7618	&0.8068	&0.6845	&0.7395	& \bf{0.7482}	& \bf{0.785}   \\
\hline
\multirow{3}{*}{SQuAD} & Magnitude
&0.4504	&0.1	&0.5816	&0.1951	&0.3318	& \bf{0.641}&0.7189		&0.1501	&0.6135	&0.0976	&0.3950	& \underline{0.588}  \\
 & Wanda
 &0.4648	&0.6686	&0.6284	&0.3536	& \underline{0.5288}	&0.611  &0.7315	&0.4291	&0.6573	&0.3170	& \underline{0.5337}	&0.533\\
 \rowcolor{gray!10} & FAPM (Ours) &0.4738	&0.7310	&0.6455	&0.5748	& \bf{0.6063}	& \underline{0.637}  &0.7410	&0.8006	&0.6752	&0.7500	& \bf{0.7417}	& \bf{0.615}   \\
  \hline
\multirow{3}{*}{MathQA} & Magnitude & 0.3812 & - & 0.4774 & 0.4314 & 0.4300 & \bf{0.818} & 0.5188 & - & 0.4361 & 0.4828 & 0.4792 & \underline{0.833}\\
 & Wanda   & 0.4417 & - & 0.6522 & 0.5625 & \underline{0.5521} & 0.800 & 0.7477 & - & 0.6714 & 0.7588 & \underline{0.7259} & 0.815 \\
 \rowcolor{gray!10} & FAPM (Ours)  & 0.4697 & - & 0.6447 & 0.5729 & \bf{0.5624} & \underline{0.812} & 0.7374 & - & 0.6899 & 0.7560 & \bf{0.7278} & \bf{0.851}\\
\hline
\multirow{3}{*}{MedQA} & Magnitude  & 0.3393 & 0.4111 & 0.4818 & 0.4438 & 0.4190 & 0.633  & 0.5844 & 0.6771 & 0.5554 & 0.6010 & 0.6044 & \bf{0.593} \\
 & Wanda  & 0.4626 & 0.7886 & 0.6519 & 0.5915 & 0.6236 & 0.608  & 0.7404 & 0.8114 & 0.6833 & 0.7490 & 0.7460 & 0.548\\
 \rowcolor{gray!10} & FAPM (Ours) & 0.4586 & 0.7733 & 0.6638 & 0.5731 & 0.6172 & 0.643  & 0.7346 & 0.8241 & 0.6935 & 0.7412 & \bf{0.7484} & \underline{0.590}   \\
 \hline
\multirow{3}{*}{MRPC} & Magnitude &0.3801	&0.6100	&0.3378	&0.4731	&0.4502	& \bf{0.892} &0.7412	&0.1296	&0.2473	&0.1768	&0.3238	& \bf{0.911}\\
 & Wanda &0.4635	&0.7845	&0.6506	&0.5958	& \bf{0.6236}	&0.816 &0.7458	&0.7989	&0.6813	&0.7482	&0.7435	&0.826  \\
 \rowcolor{gray!10} & FAPM (Ours) &0.4662	&0.7711	&0.6410	&0.5791	&0.6144	&0.882 &0.7564	&0.7938	&0.6837	&0.7682	& \bf{0.7505}	&0.892 \\
\hline
\multirow{3}{*}{QASC} & Magnitude &0.4916	&0.7263	&0.6053	&0.5223	&0.5864	&0.861&0.7559	&0.7760	&0.6407	&0.7073	&0.7199	&0.851  \\
 & Wanda &0.4705	&0.7819	&0.6456	&0.5886	&0.6216	&0.839 &0.7567	&0.8072	&0.6858	&0.7378	&0.7468	&0.828 \\
 \rowcolor{gray!10} & FAPM (Ours) &0.4836	&0.7983	&0.6326	&0.5914	& \bf{0.6265}	& \bf{0.863}  &0.7679	&0.8157	&0.6815	&0.7500	& \bf{0.7538}	& \bf{0.851}   \\
\bottomrule
\end{tabular}}
\caption{The results of FAPM with different pruning methods on various datasets on Llama3-8B and Qwen2-7B. The bold and underlined data represent the best and second-best results, respectively.}
\label{table3}
\end{table*}

In this section, we explore the performance of FAPM under different sparsity ratios. 
Figure \ref{fig5} and \ref{fig6} show the impact of FAPM on downstream task accuracy and catastrophic forgetting at different sparsity ratios on Llama3-8B and Qwen2-7B, respectively. 
As observed in Figure \ref{fig2}, using $|\Delta W|$ as the pruning criterion results in severe catastrophic forgetting at an 85\% sparsity ratio. 
However, with the application of FAPM, catastrophic forgetting is substantially mitigated even at the 85\% sparsity level. 
Notably, FAPM continues to alleviate catastrophic forgetting to some extent at a 55\% sparsity ratio in the QASC and RTE datasets, highlighting its effectiveness in preventing catastrophic forgetting.
Moreover, it was observed that downstream task accuracy significantly declines when the sparsity ratio exceeds 90\%. Conversely, when the sparsity ratio is maintained below 90\%, the impact on downstream task accuracy is minimal, although the incidence of catastrophic forgetting gradually increases. These observations suggest that a 90\% sparsity ratio may represent an optimal balance, preserving downstream task accuracy while minimizing catastrophic forgetting.

\section{Comparative Analysis of Different Pruning Criteria}
\label{pc}

One question that needs to be analyzed is why FAPM is improved based on Magnitude Pruning instead of the SOTA LLM pruning method.
We examine the application of a straightforward and efficient pruning method, 
Wanda~\citep{wanda}, to mitigate the issue of catastrophic forgetting.
Wanda addresses pruning by removing weights with the smallest magnitudes, as determined by the product of the weight magnitudes and the norms of the corresponding input activations, thereby preventing the need for retraining or weight updates, which is formulated as $S_{ij}\  =\  \left\vert W_{ij} \right\vert \cdot \| X_{j}\|_{2}$.
We prune $\Delta W$ according to this criterion in our experiments.

Table \ref{table3} presents the comparative results of FAPM and different pruning criteria methods, with all results using a 90\% sparsity ratio. 
The results reveal that while Wanda can somewhat mitigate catastrophic forgetting, it significantly impairs performance on downstream tasks. 
For example, on Llama3-8B, Wanda results in an average performance decline of 3.6\% across eight downstream datasets when compared to Full SFT, whereas Magnitude Pruning exhibits negligible impact on downstream task accuracy.
Given the necessity to preserve downstream task accuracy, we opted to use Magnitude Pruning as our foundational pruning criterion. 
Furthermore, Wanda requires a small amount of calibration data while Magnitude Pruning does not necessitate any auxiliary data.
This further reinforces our decision to select Magnitude Pruning as the basis for our method.

\section{More Analysis Results}
\label{appendix3}

\begin{figure*}[h]
\centering
  \includegraphics[width=1.0\linewidth]{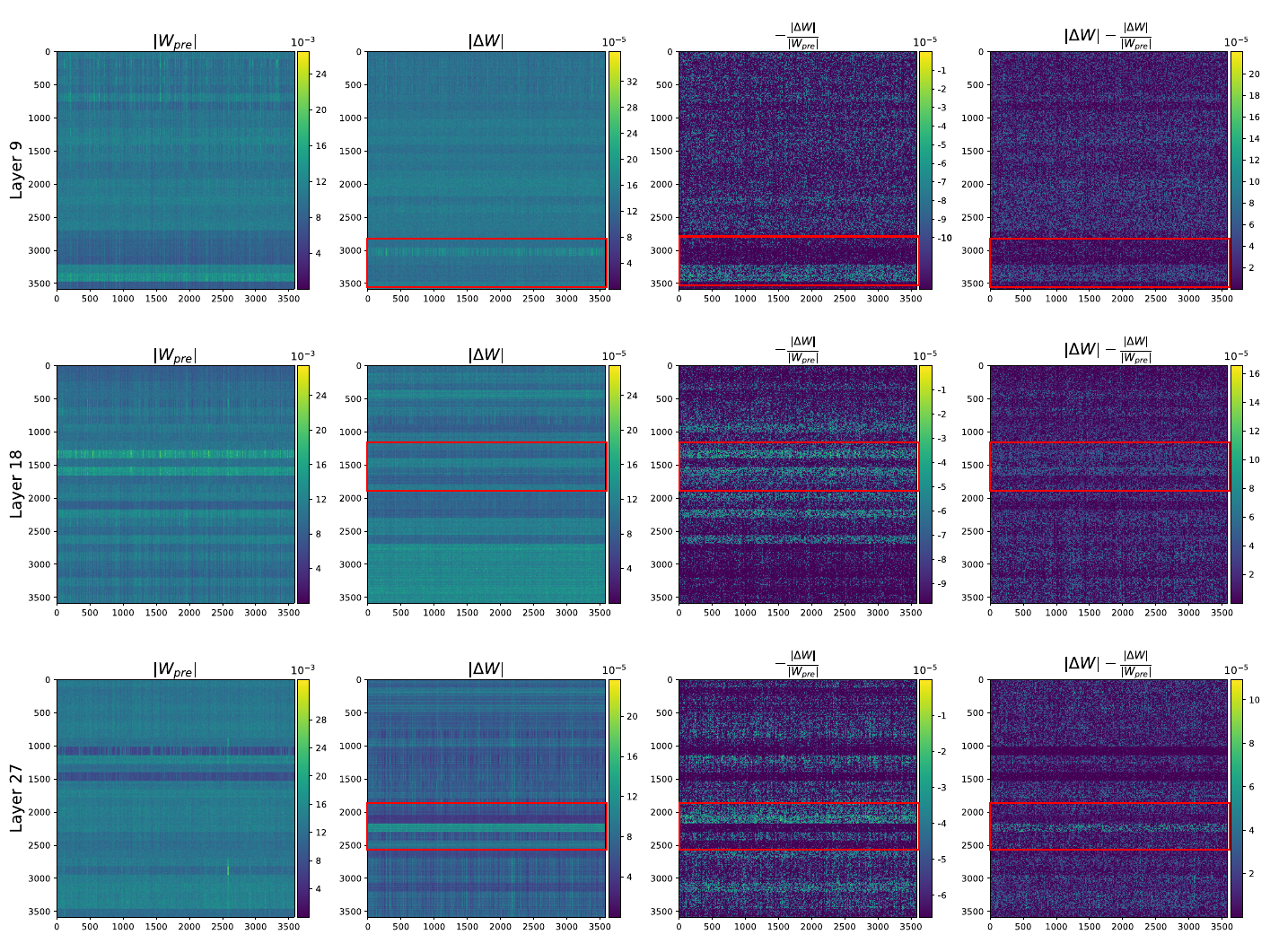}
  \caption {
   Visualization of the weight matrices in different layers of Qwen2-7B fine-tuned on RTE dataset.
   From left to right, they represent the magnitude of the pre-trained model weights, the absolute change magnitude of model weights, the relative change magnitude of model weights, and a combination of the absolute and relative change magnitude. 
}
\end{figure*}

\begin{figure}[t]
    \centering
    \subfigure[The original accuracy on RTE is 0.890 and the original average accuracy on  general tasks is 0.7556.]{
        \includegraphics[width=.40\textwidth]{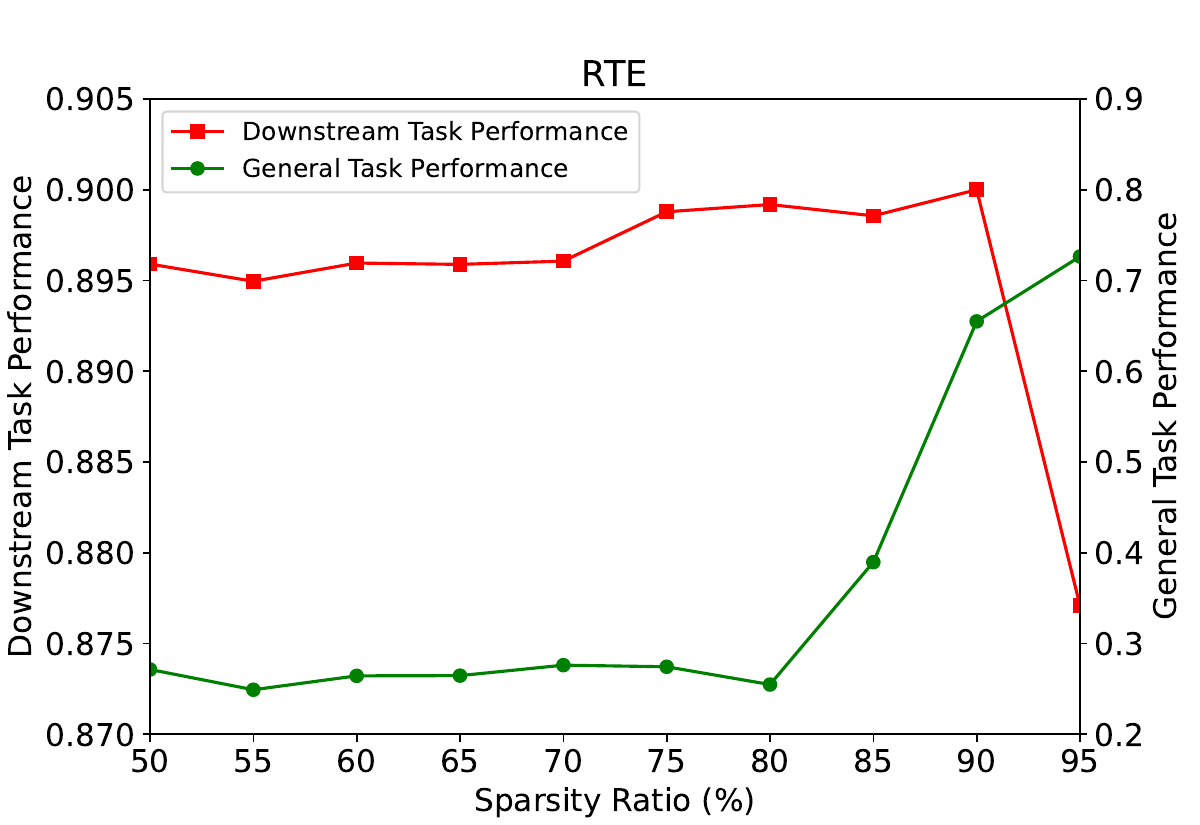}
        }
    \hspace{1cm}
    \subfigure[ The original accuracy on MRPC is 0.914 and the original average accuracy on  general tasks is 0.7556.]{
        \includegraphics[width=.40\textwidth]{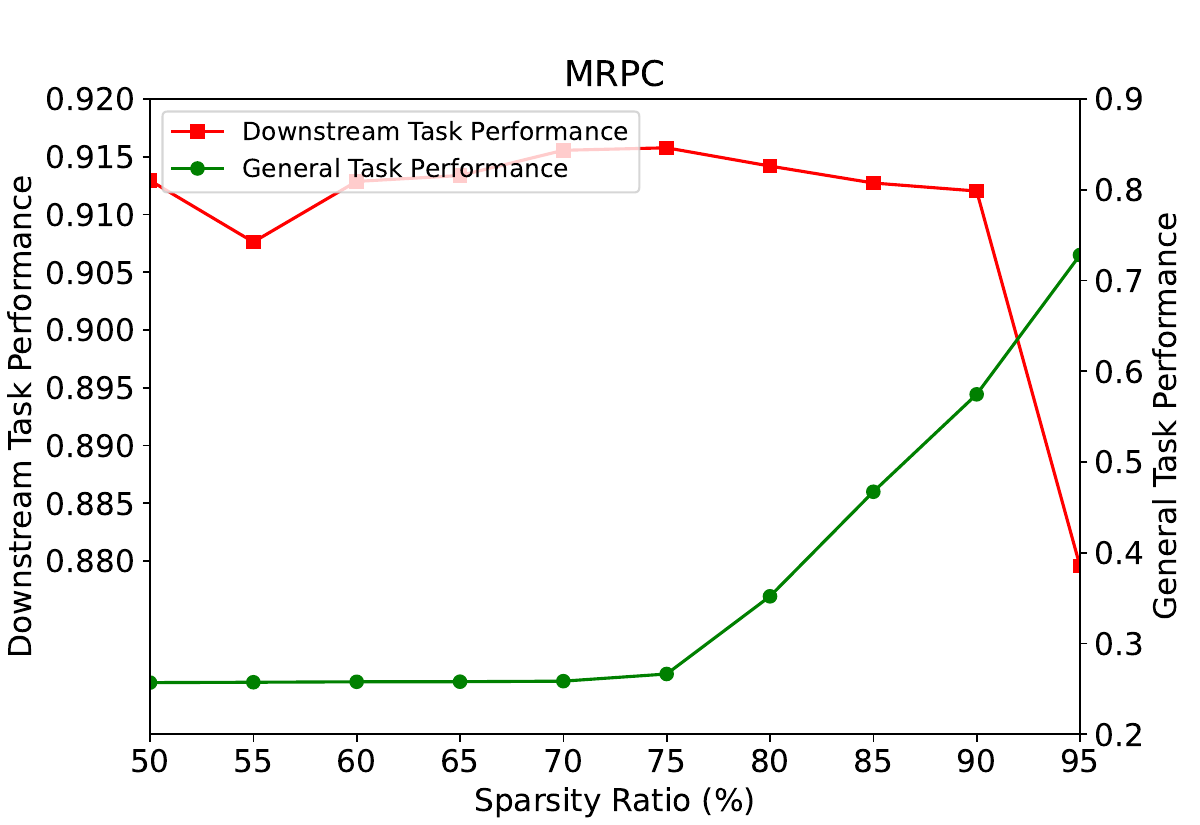}
    }
    \caption {
    The relationship between the magnitude pruning sparsity ratio, general capability, and downstream task performance of Qwen2-7B on (a) RTE and (b) MRPC, respectively.  
}
\end{figure}

\end{document}